\documentclass{article}

     \usepackage[final]{neurips_2019}

\usepackage[utf8]{inputenc} % allow utf-8 input
\usepackage[T1]{fontenc}    % use 8-bit T1 fonts
\usepackage{hyperref}       % hyperlinks
\usepackage{url}            % simple URL typesetting
\usepackage{booktabs}       % professional-quality tables
\usepackage{amsfonts}       % blackboard math symbols
\usepackage{amsmath}
\usepackage{nicefrac}       % compact symbols for 1/2, etc.
\usepackage{microtype}      % microtypography
\usepackage{color}
\newtheorem{define}{Define}
\usepackage{multirow}
\usepackage{wrapfig,lipsum,booktabs}
\newtheorem{theorem}{Theorem}
\newtheorem{lemma}{Lemma}

\usepackage{graphicx}
\setcitestyle{square,sort,comma,numbers,}

\newcommand{\nosection}[1]{\vspace{2pt}\noindent\textbf{#1.}}

\title{Generalization in Generative Adversarial Networks: A Novel Perspective from Privacy Protection}

\author{
	Bingzhe Wu$^{1,2}$,~~Shiwan Zhao$^{3,*}$, ChaoChao Chen$^{2}\thanks{Co-second author}$,~~Haoyang Xu$^{1}$
	\\
	\textbf{Li Wang}$^{2}$, \textbf{Xiaolu Zhang}$^{2}$, \textbf{Guangyu Sun}$^{1}$, \textbf{Jun Zhou}$^{2}$\\
	$^{1}$Peking University, $^{2}$Ant Financial, $^{3}$IBM Research\\
	\{wubingzhe, xuhaoyang, gsun\}@pku.edu.cn \\
	\{zhaosw\}@cn.ibm.com \\
	\{chaochao.ccc, aymond.wangl, yueyin.zxl, jun.zhoujun\}@antfin.com
}

\begin{document}

\maketitle

\begin{abstract}
In this paper, we aim to understand the generalization properties of generative adversarial networks~(GANs) from a new perspective of
privacy protection. Theoretically, we prove that a differentially private learning algorithm used for training the GAN does not overfit to a certain degree, i.e., the generalization gap can be bounded. Moreover, some recent works, such as the
Bayesian GAN, can be re-interpreted based on our theoretical insight from privacy protection. Quantitatively, to evaluate the information leakage
of well-trained GAN models, we perform various membership attacks on these models. The results show that previous Lipschitz regularization techniques are effective
in not only reducing the generalization gap but also alleviating the information leakage of the training dataset. 
\end{abstract}
\section{Introduction}
\label{sec:intro}
In the past years, generative adversarial networks~(GANs)~\cite{original_gan} have achieved
remarkable progress in a wide range of applications including image translation~\cite{cycleGAN,style-gan},
image manipulation~\cite{eccv_mainpulation, cvpr18_cartoon}, and image super-resolution \cite{srpgan, srgan}, etc. More recently, numerous advanced techniques~\cite{wgan, least_square_gan, miyato2018spectral, self-attention} are proposed for improving and stabilizing the training of GANs, leading to more realistic generated images. 

Despite the tremendous success, there are still numerous open problems to be fully solved, ranging from the theoretical analysis of different regularization techniques~\cite{miyato2018spectral, improve_gan} to the visualization of different objective functions of GANs~\cite{vis_gan}. 
%For example, how do different regularization techniques improve the training of GAN? Is there a golden-standard metric to measure the performance of a GAN model? How can we depict the landscapes of different objective functions of GAN visually?
Among these problems, a critical one is how to formally characterize the generalization ability of GANs. Some recent studies attempted to explore this problem in different contexts. For instance, a seminal work in this direction~\cite{gen_icml17} proposed the neural net distance, and the authors further showed
the generalization properties of this distance. Qi~et al. \cite{lsgan} were motivated by the progress in the Lipschitz regularization and proposed a loss-sensitive GAN.
They then developed the Lipschitz regularization theory to analyze the generalization ability of the loss-sensitive GAN. 

Different from the prior works~\cite{lsgan, gen_icml17}, in this paper, we aim to study the generalization ability of GANs in a relatively general setting from a novel perspective of privacy protection. 
Our study is motivated by a well-known intuition~\cite{overfitting_privacy}, \emph{``reducing the generalization gap''} and \emph{``protecting the individual's privacy''} share the same goal of encouraging a neural network to learn the population's features instead of memorizing the features of each individual, i.e., the smaller the generalization gap is,
the less information of the training dataset will be revealed. The goal of this paper is to validate this natural intuition
theoretically and quantitatively. 

In the theoretical side, we leverage the stability-based theory~\cite{shwartz_stable} to bridge the gap between differential privacy~\cite{algo_fund_dp} 
and the generalization, i.e., a differentially private learning algorithm 
%(used for training the discriminator) 
does not overfit to a certain degree. Based on the theoretical analysis, we also provide a new perspective from privacy protection to understand a number of recent techniques for improving the performance of GANs, e.g., various
Lipschitz regularization terms~\cite{wgan-gp, miyato2018spectral} and training GANs using Bayesian learning~\cite{bayesian_gan_nips17, he2018bayesian}. 

In the experimental side, we quantitatively validate the relationship
between the generalization gap and the information leakage of the training dataset. To this end, we introduce the \emph{membership attack} \cite{s_and_p_attack} to evaluate the information leakage of a trained GAN model. In the context of machine learning, the membership attack refers to inferring whether a specific item is from the training dataset while given the trained
model (discriminator and generator in our case). Specifically, we design different attack models and perform membership attacks on  GANs trained with various objective functions and regularization terms. The results show that previous Lipschitz regularization techniques are effective
in not only reducing the generalization gap but also alleviating the information leakage of the training dataset, which implicitly validates the aforementioned
intuition. The results also suggest that it is possible to design new variants of GAN  from the perspective of building privacy-preserving learning algorithms, which
can bring significant regularization effects while provide appealing property of protecting the sensitive information contained in the training dataset.

The rest of this paper is organized as follows. We first briefly review related works in Section~\ref{sec:related_work}. Then we
demonstrate our theoretical analysis in Section~\ref{sec:theo}. Subsequently, we present the quantitative analysis in Section~\ref{sec:exp}.
At last, we conclude this work in Section~\ref{sec:con}.

%\vspace{-4pt}
\section{Related Work}
\label{sec:related_work}
\vspace{-2pt}
\nosection{Generative adversarial networks}
In the past years, starting from the illuminative work of GANs~\cite{original_gan}, many efforts have been devoted to this research area.
Numerous researchers have tried to improve the performance of GANs from different perspectives~\cite{wgan,lsgan,fgan,miyato2018spectral}. One direction
is to improve the original objective function~\cite{lsgan, wgan, fgan}. For example, to solve the gradient vanishing problem, the Least-square
GAN~\cite{least_square_gan} proposed using the least square loss function in the training of GANs. Wasserstein GAN~(WGAN)~\cite{wgan} replaced the original Jensen–Shannon (JS) divergence with
the Wasserstein distance and proposed an effective approximation method to compute the distance. 
Besides the improvements
on the objective function, lots of algorithmic tricks have been proposed in empirical studies of training GANs. A typical direction is to add
Lipschitz constraints on the discriminator, which enables the discriminator to be Lipschitz continuous with respect to the input and the weight.
For instance, WGAN proposed using weight clipping to constrain the Lipschitz constant~\cite{wgan}. Gulrajani~et al.~\cite{wgan-gp} further suggested to use the gradient penalty to obtain better performance. Miyato~et al. \cite{miyato2018spectral} took a different angle to regularize
the spectra of the weight matrix which can implicitly constrain the Lipschitz constant.  

Among these empirical techniques in training GANs, some researchers focus on building the theoretical framework to analyze the properties of GANs
under different assumptions. Here we focus on the works of studying the generalization properties of GANs. Specifically, Arora et al.~\cite{gen_icml17} argued that
the objective function of the original GAN does not provide a theoretical generalization guarantee. As a result, the authors turned into analyzing the generalization gap of their
proposed neural network distance. Qi et al. \cite{lsgan} developed a set of theoretical tools to analyze the generalization of GANs under the assumption of Lipschitz continuous of the discriminator. In practical, to meet the assumption, they designed a novel objective function to directly minimize the Lipschitz
constant.

\nosection{Membership attacks towards deep learning algorithms} Recently, membership attacks have arisen as a common threat model against machine learning algorithms and attained increasing attraction from the research community \cite{attack_18stacey,s_and_p_attack, secret_sharer, gan_membership}. 
A pioneer work~\cite{s_and_p_attack} investigated the risk of membership attacks on different machine learning models. Specifically, they developed a shadow training technique to obtain an attack model in the black-box setting (i.e., without knowing the machine learning model structure and parameters). Carlini et al.
\cite{secret_sharer} proposed a metric to measure the vulnerability of deep learning models. In the context of GANs, Hayes et al.~\cite{gan_membership} studied membership
attacks against GANs in both black-box and white-box settings. 
In this paper, we use the membership attack to assess the information leakage of the dataset used for training GAN models.
\section{Theoretical Analysis}
\label{sec:theo}
In this part, we aim to bridge the gap between the privacy protection and the generalization
ability of GANs. At a high level, we prove that a differentially private
learning mechanism~(i.e. training algorithm) does not overfit
to a certain extent. The core idea is based on
the stability-based theory~\cite{shwartz_stable}. %To this end, we will first introduce some basic notations in GAN. Then we can present the main theoretical analysis to understand the generalization of GAN from the perspective of privacy protection. At last, based this new insight, we can show a new interpretation of the previous theoretical results~(uniform convergence)~\cite{gen_icml17}. And we can build the connection between our perspective and the recent efforts on Bayesian GAN.
To this end, we first introduce some basic notations in GANs. Then we present the theoretical analysis to characterize the generalization ability of GANs from the perspective of privacy protection. At last, based on our theoretical insight from privacy protection, we show a new interpretation of the previous theoretical results of uniform convergence~\cite{gen_icml17}, as well as the recent efforts on the Bayesian GAN.

\nosection{Preliminaries} The GAN framework consists of a generator and a discriminator. We denote ${\mathcal{H}_{G}}$ and ${\mathcal{H}_{D}}$ as the hypothesis
spaces of the generator and discriminator, respectively. In practice, we make use of neural
networks to build $\mathcal{H}_G$ and $\mathcal{H}_D$. Formally, we have $\mathcal{H}_G = \{\mathbf{g}(z;\theta_g), \theta_g \in \mathbb{R}^p \}$ and
$\mathcal{H}_D = \{\mathbf{d}(x;\theta_d), \theta_d \in \mathbb{R}^q \}$, where $\mathbf{g}$ and 
$\mathbf{d}$ are multi-layer convolutional neural networks. $\theta_g$ and $\theta_d$ are the corresponding weight parameters. The training of GANs can be seen as
playing a min-max game to solve the following
optimization problem:
\begin{equation}
    \min_{\theta_g}\max_{\theta_d} \mathbb{E}_{x\sim p_{data}}[\phi(\mathbf{d}(x;\theta_d))]+\mathbb{E}_{z\sim p_{z}}[\phi(1-\mathbf{d}(\mathbf{g}(z;\theta_g);\theta_d))]
\label{eq:loss_obj}
\end{equation}
The above formulation can be seen as an extended version of the objective function 
used in the vanilla GAN~\cite{original_gan}. According to previous literature~\cite{gen_icml17}, we call function $\phi$ the
\emph{measuring function}. 
Note that setting $\phi(t)=\log(t)$ leads to the objective function used in the work of the
original GAN~\cite{original_gan}, while the recent WGAN~\cite{wgan} proposed using $\phi(t) = t$.

To optimize Equation \ref{eq:loss_obj}, we need to build two learning mechanisms $\mathcal{A}_d$ and $\mathcal{A}_g$. During the training, we alternately run $\mathcal{A}_d$ and $\mathcal{A}_g$ to seek an equilibrium $(\theta_d^*, \theta_g^*)$. Specifically, $\mathcal{A}_d$
tries to find $\theta_d^*$ such that it maximizes the expected loss of the discriminator:
\begin{equation}
   U(\theta_d, \theta_g^*) = \mathbb{E}_{x\sim p_{data}}[\phi(\mathbf{d}(x;\theta_d))]+\mathbb{E}_{z\sim p_{z}}[\phi(1-\mathbf{d}(\mathbf{g}(z;\theta_g^*);\theta_d))]
   \label{eq:loss_d}
\end{equation}
and $\mathcal{A}_g$ tries to find $\theta_g^*$ to minimize the expected loss of the generator:
\begin{equation}
    V(\theta_d^*, \theta_g) = \mathbb{E}_{z\sim p_{z}}[\phi(1-\mathbf{d}(\mathbf{g}(z;\theta_g);\theta_d^*))]
    \label{eq:loss_g}
\end{equation}

\nosection{Empirical loss and generalization} The optimization of Equation \ref{eq:loss_d} and \ref{eq:loss_g} can not be directly solved
since the expectation over the distribution of the true data $p_{data}$ is intractable. Instead,
we approximate them with empirical loss on a set of i.i.d. real data samples $S = \{x_1, x_2,  \cdots, x_m\}$ and noise vectors $Z = \{z_1, z_2, \cdots, z_m\}$ drawn from $p_{data}$ and $p_{z}$, respectively\footnote{In practice, we always sample different $Z$ while $S$ is fixed
at each training round.}. 
%We formally denote the sampling process as $S\sim \mathcal{D}^m$, where $S=X\times Z$ and $\mathcal{D}$ is the
%joint distribution of the noisy and the real data. The superscript $m$ denotes the number of samples.  
We denote the resulted empirical versions of Equation \ref{eq:loss_d} and \ref{eq:loss_g} as $\hat{U}$ and $\hat{V}$. In this empirical setting, the learning mechanisms
$\mathcal{A}_d$ and $\mathcal{A}_g$ turn into the role as the empirical loss optimizers, which are to optimize the empirical loss $\hat{U}$ and $\hat{V}$, i.e., finding $(\hat{\theta}_d^*, \hat{\theta}_g^*)$. To study the generalization property of the learning mechanisms, we need to evaluate the generalization gap between the empirical and expected objective losses.
In this paper, we mainly focus on the analysis of Equation \ref{eq:loss_d}
since our viewpoint is from the privacy protection and Equation \ref{eq:loss_g} does not explicitly touch the original training data. As a common practice shown in the prior
work~\cite{original_gan}, we analyze Equation \ref{eq:loss_d} when $\theta_g^*$ is given. 
% Note that we can infinitely sample from $p_z$(\emph{e.g.} sampling from a uniform distribution) and $p_z$ is irrelevant to the original training data. As a result, we omit the second term in the RHS of Equation \ref{eq:loss_d} and focus on studying the first term.
Formally, we define the generalization gap as follows~(we take the discriminator loss $U$ as an example):
%\begin{equation}
%    F_{U}(\mathcal{A}_d) = \mathbb{E}_{\mathcal{A}}\mathbb{E}_{S\sim p^m_{data}}[\hat{U}(\mathcal{A}_d(S), \theta_g^*)-U(\mathcal{A}_d(S), \theta_g^*)]
%    \label{eq:gen_err}
%\end{equation}
\begin{equation}
    F_{U}(\mathcal{A}_d) = \mathbb{E}_{\theta_d\sim\mathcal{A}_d(S)}\mathbb{E}_{S\sim p^m_{data}}[\hat{U}(\theta_d, \theta_g^*)-U(\theta_d, \theta_g^*)]
    \label{eq:gen_err}
\end{equation}
where $S\sim p^m_{data}$ denotes sampling $m$ training samples from the oracle distribution $p_{data}$.
In the above equation, we take the expectation with respect
to the randomness in the learning mechanism and also in the sampling process similar to the literature~\cite{sgld,yuxiang_dp_stable}. Note that we can infinitely sample from $p_z$ (e.g. a uniform distribution) and $p_z$ is irrelevant to the original training data (i.e. sampling from $p_z$ does not induce the leakage of the training dataset). 
As a result, we omit the second term in the RHS of Equation \ref{eq:loss_d} and focus on studying the first term.

\nosection{Privacy protection and generalization bound}
To bridge the gap between privacy protection and the generalization bound, we need to characterize
how an algorithm can protect privacy, i.e., the amount of information leakage of the training dataset. Differential privacy~\cite{dwork2011differential}
is seen as a gold standard for privacy
protection in the security community. It provides a rigorous bound on privacy cost of the algorithm, even in the worst case.
The definition of differential privacy is based on the adjacent datasets. Two datasets are adjacent when they differ in a single element.
Then we can introduce differential privacy as follows:
\begin{define}
\label{definition:dp}
\textbf{(Differential privacy)} A randomized algorithm $\mathcal{A}: D\to~R$ satisfies
$\epsilon$-differential privacy if for any two adjacent datasets $\mathcal{S}, \mathcal{S}^{'}\subseteq D$ and for any subset of outputs $O \subseteq R$ it holds:

\begin{equation}
    \mathbf{Pr}[\mathcal{A}(\mathcal{S}) \in O] \leq e^{\epsilon} \mathbf{Pr}[\mathcal{A}(\mathcal{S}^{'}) \in O]
    \label{eq:dp}
\end{equation}
\end{define}
In our setting, $\mathcal{A}$ can be the training algorithm (i.e. $\mathcal{A}_d$). Intuitively, Equation~\ref{eq:dp}
indicates that participation of one individual sample in the training phase has a negligible effect on the final weight parameters. 
A relevant concept is \emph{uniform RO-stable} of an algorithm. An algorithm is stable if a small change to the input
causes a limited change in the output. Here, RO denotes \emph{``replace one element in the input``}. The above description
is made formally as:
\begin{define}
\label{define:stable}
\textbf{(Uniform RO-stability)} The randomized algorithm $\mathcal{A}$ is uniform RO-stable with respect to the discriminator loss function (Equation~\ref{eq:loss_d}) in our case, if for
all adjacent datasets $S, S'$, it holds that:
\begin{equation}
    \sup_{x\in S}|\mathbb{E}_{\theta_d\sim\mathcal{A}(S)}[\phi(\mathbf{d}(x;\theta_d))]-\mathbb{E}_{\theta_d\sim\mathcal{A}(S')}[\phi(\mathbf{d}(x;\theta_d))]|\leq \epsilon_{stable}(m)
\end{equation}
\end{define}
A well-known heuristic observation is
that differential privacy implies uniform stability. The
prior work~\cite{yuxiang_dp_stable} has formlized this observation into the
following lemma:
\begin{lemma}
(\textbf{Differential privacy $\Rightarrow$ uniform RO-stability}) If a randomized algorithm $\mathcal{A}$ is $\epsilon$-differentially
private, then the  algorithm $\mathcal{A}$ satisfies ($e^{\epsilon}-1$)-RO-stability.
\end{lemma}
%Despite the relation to differential privacy, 
The stability of the algorithm is also related to the generalization gap. Numerous studies~\cite{shwartz_stable, sgld}
focus on exploring the relationship in various settings. Formally, we have the following lemma:
\begin{lemma}
If an algorithm $\mathcal{A}$ is uniform RO-stable with
rate $\epsilon_{stable}(m)$, then $|F_U(\mathcal{A})|$ (Equation~\ref{eq:gen_err}) can
be bounded: $|F_U(\mathcal{A})| \leq \epsilon_{stable}(m)$.
\end{lemma}
Intuitively, the more stable the algorithm is, the better
its generalization ability will be. We take a further step 
to build the connection between differential privacy
and the generalization gap. This can be done via
combining the above two lemmas. Formally, we introduce
Theorem~\ref{theo:gen_gap} as follows:
\begin{theorem}
(\textbf{Generalization gap}) If an algorithm $\mathcal{A}$ satisfies $\epsilon$-differential privacy, then the generalization gap can be bounded by
a data-independent constant.
\label{theo:gen_gap}
\end{theorem}
The proof can be accomplished
by following the roadmap: $Differential~privacy \Rightarrow Stability  \Rightarrow Generalization$. The proof details can be found in Appendix.
By applying Theorem~\ref{theo:gen_gap} to $\mathcal{A}_d$, we can show that the generalizability of the discriminator
is ensured when the training algorithm satisfies $\epsilon$-differential privacy. Note that we focus on the generalization of the discriminator loss, since the optimization of Equation~\ref{eq:loss_g} does not touch the original 
data. We can easily obtain the similar generalization bound of the generator by leveraging the post-processing  property of the differential privacy protocol~\cite{algo_fund_dp}.

Theorem~\ref{theo:gen_gap} not only 
enables characterizing GAN's generalizability from the viewpoint of privacy protection, but also helps to understand
previous techniques for improving the generalization of GANs.
A typical example is the Lipschitz regularization technique. Previous studies propose implementing the regularization from
various angles~\cite{wgan,lsgan,miyato2018spectral}. For instance, the loss-sensitive GAN~\cite{lsgan} designed a novel objective loss to restrict the discriminator to satisfy the Lipschitz condition.
Spectral normalization~\cite{miyato2018spectral} explored this direction by adding regularization on the weight parameters. And WGAN~\cite{wgan} proposed using gradient penalty
to constrain the magnitude of the gradient, which further implicitly led to the Lipschitz condition. From the perspective of differential privacy, 
the Lipschitz condition for the outputs is also a crucial ingredient for building a differentially private algorithm. We also infer that adding Lipschitz constraints
implicitly leads to stability of the algorithm, which can further be used for reducing the generalization gap (see more details in the evaluation section).

Above analysis focuses on the discriminator loss $U$. As mentioned above,  it is natural to extend the analysis to the generator loss 
since the optimization of the generator loss does not touch the original training dataset and we can leverage the post-processing property of the differential privacy. We can further study the whole optimization procedure of the GAN, i.e., the alternative running of $\mathcal{A}_d$ and $\mathcal{A}_g$. These can be accomplished by the composition theory in adaptive learning theory~\cite{AdaLearning}, we mark this as the future work.

\nosection{Revisiting previous results on uniform convergence} Previous works attempted to explore the uniform convergence with respect to different objective functions. For example, Qi et al.~\cite{lsgan} proposed the loss-sensitive GAN and proved the uniform convergence with respect to the discriminator loss.
Arora et al. proposed the neural distance to analyze the generalization property (uniform convergence) of GANs. Note that both of them focus on the special form of the GAN or the objective function. In this paper, based on the aforementioned
theoretical results, we can prove the uniform convergence of the discriminator loss when the training algorithm satisfies the differential privacy protocol. Formally,
we have the following theorem:
\begin{theorem}
(\textbf{Uniform convergence}): Suppose $\mathcal{A}_d$ satisfies $\epsilon$-differential privacy and $\mathbf{d}^{(k)}(x;\theta_{d}^{(k)})$ be the output of $\mathcal{A}_d$ at the $k$-th iteration. Then, $\forall k$, the generalization gap with respect to $\mathbf{d}^{k}$ can be bounded by a universal constant which
is related to $\epsilon$.
\end{theorem} 

The proof of the above theorem can be done via combing the post-processing property \cite{algo_fund_dp} of differential privacy and McDiarmid's inequality \cite{vershynin2018high}. The details can be found
in Appendix.

\nosection{Connection to Bayesian GAN}
Recently, training GANs using Bayesian learning has emerged as a new way to avoid mode collapse \cite{bayesian_gan_nips17}. 
A well-known interpretation of mode
collapse is that the generator/discriminator has
memorized some examples from the training dataset.
Hence, the memorization
phenomenon can also breach the privacy of individuals in the training dataset. Thus, we infer that the effectiveness of the Bayesian GAN may come from preventing information leakage of the training dataset.
In what follows, we briefly introduce how we validate this conjecture with our theoretical results. 
Specifically, we take a recent
work~\cite{bayesian_gan_nips17} as an example. In the work,
the authors proposed using stochastic Hamiltonian 
Monte Carlo (HMC) to marginalize the posterior distribution of the weight parameters of both generator and
discriminator. We have noted that Wang et al. \cite{privacy_for_free} pointed out that sampling one element from
a posterior distribution can implicitly satisfy differential privacy. Based on their results, we can
prove that the HMC sampling also preserves differential privacy with minor modification (refers to Section 4 in \cite{privacy_for_free}), and then the Bayesian GAN can implicitly preserve the differential privacy.
Thus we are not surprised
that the Bayesian GAN can alleviate mode collapse in GANs since its connection to differential privacy mentioned above.

%\vspace{-0.3cm}
\section{Quantitative Analysis}
% \usepackage{multirow}

\label{sec:exp}
\vspace{-0.1cm}
In this section, we quantitatively validate the connection between the generalizability and the privacy protection/information leakage of GANs by investigating some of the most popular GAN models. In the theoretical analysis, we focus on the differentially private learning algorithms. While in practice, differential privacy is a strict requirement for most of the existing GANs so that we focus on studying information leakage of GANs instead. 
In particular, we note that adding Lipschitz constraints on the discriminator has recently emerged as
an effective solution for improving the generalization ability of such GANs, thus we aim to study the
effects of various regularization techniques for adding the Lipschitz constraints. In a nutshell, our results show that the Lipschitz constraints
not only reduce the generalization gap but also make the trained model resistant to the membership attacks which are used for detecting the
sensitive information contained in the training dataset. 

This section is structured as follows. First, we will introduce the experimental settings including the datasets and the choice of different hyper-parameters. Then, we demonstrate our main results on different datasets.
At last, we provide some discussions on the attack methods and the regularization techniques.
\vspace{-0.1cm}
\subsection{Experimental Setup}
\vspace{-0.2cm}
\nosection{Datasets} We conduct experiments on a face image dataset and a real clinical dataset, namely, Labeled Faces in the Wild~(LFW)~\cite{LFWTechUpdate} which consists of 13,233 face images,  and the IDC dataset which is publicly available for invasive ductal carcinoma (IDC) classification\footnote{{\tt http://www.andrewjanowczyk.com/use-case-6-invasive-ductal-carcinoma-idc-segmentation/}} and contains $277\rm{,}524$ patches of $50\times50$ pixels ($198\rm{,}738$ IDC-negative and $78\rm{,}786$ IDC-positive).

\nosection{Model setup}
Note that we focus on studying the effects of different regularization techniques instead of the architecture design of
the GAN model, thus we use the same generator architecture and the same discriminator architecture in all experiments. Specifically, we adopt DCGAN following most of previous works~\cite{dcgan}. More details can be found in the work~\cite{dcgan}. We set the size of the generated image to 64x64 for the LFW dataset and 
32x32 for the IDC dataset, respectively. As
for optimization, we use Adam~\cite{adam} in all experiments, and use different hyper-parameters for different training strategies. To be specific, we make use of Adam for the GAN trained with JS divergence. 
%The learning rates is set to $0.0002$ except for the GAN trained without any regularization terms (original GAN~\cite{original_gan}). For the original GAN, we set the learning rate to $0.0004$. For GANs trained using Wasserstein distance, the learning rate is set to $0.0002$. 
The learning rates is set to $0.0004$ for the GAN trained without any regularization terms (original GAN~\cite{original_gan}), while for other GANs (e.g. trained using Wasserstein distance), the learning rate is set to $0.0002$. 
More details of hyper-parameter settings (e.g. $\beta$ in Adam)  can be found in Appendix. We trained all the models for
400 epochs on both datasets. 

\nosection{Attack setup}
We make use of membership attacks for evaluating the information leakage of the GAN model.  
We build the attack model based on the output of the discriminator, which is a bounded function. We suppose $\mathbf{d}(x;\theta_d)\leq b$ for all $x$ in Equation~\ref{eq:loss_obj}. This assumption naturally holds
in the context of GAN (e.g. $b=1$ for the original GAN). This is also a common assumption in many previous works. Here, letting $b=1$ suffices to all our cases. We then assume that the attacker $\mathcal{A}$ has access to the trained GAN model, i.e. the discriminator and generator. Note that the notation $\mathcal{A}$ is different from the previous ones that denote training algorithms.  The goal of the attacker is to determine
whether a record (an image in our case) in the attack testing dataset is from the original training dataset.  Based on the above setting, the attack model proceeds as follows:
\begin{itemize}
    \item Given the discriminator $\mathbf{d}(x;\theta_d)$ and an image from the attack testing dataset.
    \item $\mathcal{A}$ firstly sets a threshold $t\in (0,1)$.
    \item $\mathcal{A}$ outputs 1 if $\mathbf{d}(x;\theta_d)/b \geq t$, otherwise, it outputs 0.
\end{itemize}
where the output of 1 indicates that the input image is from the training dataset. 

To evaluate the performance of the attack model, we need to build the
attack testing dataset. For the LFW dataset, we randomly choose 50\% of images as the original training dataset to train the GAN model. We build the attack
testing dataset by mixing the original training dataset and the remaining images. For the IDC dataset, we only use the positive part of the dataset. Since
the data provider has already provided a partition of training and testing datasets~\cite{janowczyk2016deep} ($22\rm{,}383$ images in the training dataset and $14\rm{,}157$ images in the testing dataset\footnote{They also provide a validation dataset and  we did not use it in our experiments.}), we directly use the original partition and build the attack testing
dataset by mixing the training and testing datasets. We then treat the above attack as a binary classification model and evaluate its performance based on the F1 score and the AUC\footnote{Area under the ROC curve} value.
To compute the F1 score, we assume the attacker has obtained the average value of $\mathbf{d}(x;\theta_d)$ on the training dataset. Thus we can set the average value as the threshold $t$ and then compute the F1 score at this threshold. 

\subsection{Results on LFW and IDC Datasets}
Here, we present the results of our evaluation on the GAN models trained with different strategies. We conduct extensive experiments of different settings. Specifically, we focus on three commonly used techniques for adding Lipschitz constraints, namely, weight clipping~\cite{wgan},
gradient penalty~\cite{improve_gan}, and spectral normalization~\cite{miyato2018spectral}. For weight clipping, we set the clip interval as $[-0.01, 0.01]$ following the prior work~\cite{wgan}. We combine
these regularization techniques with two types of objective functions, the traditional JS divergence~(setting $\phi(t)=\log(t)$) and the Wasserstein distance~
(setting $\phi(t)=t$). As mentioned above, the performance of the attack model is measured by the F1 score and AUC value, and we use the gap between the testing and
training losses to estimate the generalization gap. We also calculate the Inception score~\cite{improve_gan}, which is a commonly used for assessing the quality of the generated images in
previous works. 
The overall results are shown in Table~\ref{tab:overall_results}.

\begin{table}[httb]
\caption{Evaluation results of DCGAN trained with different strategies. IS denotes the Inception score. N/A indicates that the strategy leads to failure/collapse of the training. The last row presents the Inception
scores of the real data (training images of these two datasets).}
\label{tab:overall_results}
\centering
\begin{tabular}{ccccccccc}
\toprule
\multirow{2}{*}{Strategy} & \multicolumn{4}{c}{LFW}                                                         & \multicolumn{4}{c}{IDC}                                                              \\ \cline{2-9}
                          & \multicolumn{1}{c}{F1} & \multicolumn{1}{c}{AUC} & \multicolumn{1}{c}{Gap} & \multicolumn{1}{c}{IS} & \multicolumn{1}{c}{F1}    & \multicolumn{1}{c}{AUC}   & \multicolumn{1}{c}{Gap} & \multicolumn{1}{c}{IS} \\ \midrule
\textbf{-JS divergence-}  & \multicolumn{8}{l}{}                                                                                                                                                   \\ 
Original                  &\multicolumn{1}{c}{0.565}            & \multicolumn{1}{c}{0.729}             & \multicolumn{1}{c}{0.581}                     &3.067    & \multicolumn{1}{c}{0.445} & \multicolumn{1}{c}{0.531} & 0.138                   &  2.148 \\ 
Weight Clipping           & \multicolumn{1}{c}{0.486}                        & \multicolumn{1}{c}{0.501}                        &  0.113                       &3.112   & 0.378                     & 0.502                    & 0.053                   & 2.083  \\ 
Spectral Normalization    & 0.482                     & 0.506                        & 0.106                        &  3.104  & 0.416                     & 0.508                     & 0.124                   &2.207   \\ 
Gradient Penalty          &   \multicolumn{4}{c}{N/A} & \multicolumn{4}{c}{N/A}  \\ \midrule
\textbf{-Wasserstein-}      & \multicolumn{8}{l}{}                                                                                                                                                   \\ 
W/o clipping& \multicolumn{4}{c}{N/A} & \multicolumn{4}{c}{N/A}\\
Weight Clipping           & 0.484                        &0.512                      &  0.042                       &3.013    & 0.388                     & 0.513                     &0.045                        & 1.912   \\ 
Spectral Normalization    & 0.515                       & 0.505                        &0.017                         &3.156   &0.415                           &0.507                           &0.013                         & 2.196   \\ 
Gradient Penalty          &  0.492                     & 0.503                         &0.031                         &  2.994  &0.426                         & 0.504                          &0.017                         & 1.974   \\ \midrule
IS (Real data)            & \multicolumn{4}{c}{4.272}                                                            & \multicolumn{4}{c}{3.061}                                                                 \\ \bottomrule
\end{tabular}
\end{table}
\vspace{-3pt}
\nosection{LFW}
We first conduct contrastive experiments on the GAN trained using JS divergence (setting $\phi(x) = \log(x)$ in Equation~\ref{eq:loss_obj}). 
As shown in Table~\ref{tab:overall_results}, the plain model~(trained without any regularization term) is more
vulnerable than those trained with different regularizers. The plain model leaks some information of the training dataset, which
results in the F1 score of $0.565$ and the AUC value of $0.729$ (greater than $0.5$), respectively. While various regularization techniques are used for training the GAN,
the attack performance decreases drastically. For instance, while spectral normalization is used, the F1 score is dropped from $0.565$ to $0.482$, and the AUC value is dropped to $0.506$, which approximates to the random guess. Along with the decrease of the attack performance, we also observe a consistent decrease in the generalization gap (Gap in Table~\ref{tab:overall_results}). For example, the gap is decreased from $0.581$ to $0.106$ while spectral normalization is used. 

In addition to these metrics, we also
calculate the Inception score of each model. As shown in  Table~\ref{tab:overall_results}, the Inception score of the LFW dataset~(shown in the last row of Table~\ref{tab:overall_results}) is much
smaller than the commonly used benchmarks~(e.g. Cifar10~\cite{cifar10} and ImageNet~\cite{imagenet}). This is because that the Inception score is calculated by the model pre-trained by the images in ImageNet in which
the image always contains general objects (such as animals) while
the image in LFW always contains one face. We observe that the spectral normalization can generate images with higher visual qualities but obtain lower
Inception score than the weight clipping strategy~(see generated images in Appendix). Thus, these numerical results of Inception score indicate that Inception score is not suitable for some
image generation tasks and we need to design a specific metric for a given task. Among these experiments, we also conduct the same attack experiments on GANs trained using Wasserstein distance and observe similar phenomena as shown in Table~\ref{tab:overall_results}. 

\nosection{IDC}
The images in the IDC dataset contain various tissues and all of these tissues have similar shapes and colors. As a result, performing the attack on the IDC dataset
is more challenging than on other datasets which consist of some concrete objects (e.g. LFW and Cifar10~\cite{cifar10}). As we can see in Table~\ref{tab:overall_results},
in all cases, the attack performance on IDC is lower than the performance on LFW in terms of the F1 score and AUC value. We also provide some quantitative analysis to interpret the performance drop in the following subsection. Despite the performance drop, we can still observe the effectiveness
of different regularizers in reducing the generalization gap and information leakage. For example, with the use of spectral normalization, AUC drops from 0.531 to 0.508 and F1 score decreases from
0.445 to 0.416. Another notable thing is that training the GAN using Wasserstein distance without weight clipping can lead to the failure of training. This may be caused by the gradient explosion since there is no activation function to compress the output values of the last layer of the discriminator in Wasserstein GAN (in contrast, the original GAN used the sigmoid function for the compression purpose). For the results of Inception score, we can observe an obvious decrease from
the LFW dataset to the IDC dataset. This may be caused by the difference of contents contained in the images (tissues in IDC images and faces in LFW images).

In summary, all the empirical evidence implies that previous regularizers for adding Lipschitz constraints  can not only decrease the generalization gap but also reduce the
information leakage of the training dataset in terms of the attack performance.
Moreover,
we observe that spectral normalization achieves comparable visual quality of the generated images in contrast to the original GAN~(see generated images in Appendix). This
suggests that we can attempt to use this technique in various applications to trade off the information leakage and the visual quality of the generated images. 

\subsection{Discussions}
\vspace{-0.3cm}
\nosection{Attack performance on different datasets}
In the experiments, we observe that the attack performance may vary on different datasets. From Table~\ref{tab:overall_results}, in all cases, the attack performance
of IDC is always worse than the performance of LFW. For the original GAN model, the
AUC of the attack model on LFW is 0.729 while the one for IDC is 0.531 which shows a 27.2\% relative performance drop. We infer that the drop is caused by the higher similarity of the images in the IDC data. Quantitatively, the similarity can be measured by the standard deviations~(per channel for an RGB image) of these two datasets. 
Specifically, the standard deviation of IDC is 0.085~(R-channel) while the value of LFW is 0.249~(R-channel). The IDC dataset has significantly lower standard deviation than the LFW dataset. 
This suggests that an individual example in the IDC dataset will be less likely to noticeably impact the final model which further constrains an adversary's ability. Note
that similar evidence has been found in the prior work~\cite{s_and_p_attack}.

\nosection{Other weight normalization} In the previous experiments, we have employed three regularizers for limiting the discriminator to be
Lipschitz continuous. In addition to these regularizers, there are some other approaches to use weight normalization techniques to regularize the
model training. The \emph{original weight normalization} introduced in the work \cite{weight_norm_16} is to normalize the l2 norm of
each row vector in a weight matrix. Subsequently, Brock et al.~\cite{orth_norm} proposed \emph{orthonormal regularization} on the weight matrix to stabilize the
training of GANs. We also conduct attack experiments on the GANs trained with these two normalization methods. In practice, the orthonormal regularization achieves comparable performance in terms of the F1 score of the attack model
(0.402 for the IDC dataset), while obtains comparable image quality compared with spectral normalization. In addition, the original weight normalization will lead to training failure (not convergence/mode collapse) in our cases. The failure
may be caused by the conflict between the weight normalizing and the desire
to use as many features as possible as discussed in the prior work~\cite{miyato2018spectral}. 
\begin{wraptable}{r}{8cm}
\caption{The results of black-box attack on the LFW dataset.}
\label{tab:bbbox_attack}
\begin{tabular}{cccc}
\toprule
Strategy & F1 & AUC & Gap \\ \hline
Original &0.423   &0.549        &0.581    \\ \hline
Weight Clipping  &0.358 &0.502   & 0.113\\ \hline
Spectral Normalization &0.347   &0.497       &0.106    \\ \bottomrule
\end{tabular}
\end{wraptable}

\nosection{Black-box attack}
The aforementioned analysis is all based on the white-box attack, i.e., the model structure
and weights of the discriminator and generator are exposed to the adversary.
In this part, we present some discussions on the black-box attack. We consider
black-box attacks where the adversary has limited auxiliary knowledge
of the dataset follow the prior work~\cite{gan_membership}. Specifically,
for the LFW dataset, we assume the attacker has 30\% images of both the
training and testing datasets (marked as the auxiliary knowledge in the following part). Then we can use the auxiliary information to build a "fake" discriminator
to imitate the behavior of the original discriminator~(more details can be found in the prior work \cite{gan_membership}). Once the "fake" discriminator is obtained, we can replace the original discriminator with the fake one to perform the aforementioned
white-box attack. The results of the black-box attack are summarized in
Table~\ref{tab:bbbox_attack}. Intuitively, the black-box is a more challenging
task than the white-box attack. This can be observed by the decrease of the attack performance from the white-box to black-box~(the AUC decreases from 0.729 to 0.549 for the original model).
From Table~\ref{tab:bbbox_attack}, we can also
observe the effectiveness of the spectral normalization for reducing the information leakage.  
% \usepackage{multirow}

%\vspace{-0.3cm}
\section{Conclusion}
%\vspace{-0.3cm}
\label{sec:con}
In this paper, we have shown a new perspective of privacy protection to
understand the generalization properties of GANs. Specifically,
we have validated the relationship between the generalization gap and
the information leakage of the training dataset both theoretically and quantitatively. This new perspective can also help us to understand the
effectiveness of previous techniques on Lipschitz regularizations and the Bayesian GAN.
We hope our work can light the following researchers to leverage our new perspective to design GANs with better generalization ability while preventing the
information leakage.
\bibliographystyle{plain}
\bibliography{nips19}
\newpage
\appendix
\section{Appendix}
\subsection{Proof of Theorem 1}
Before presenting the proof of Theorem 1, we first re-demonstrate the previous
lemmas here.
\begin{lemma}
(\textbf{Differential privacy $\Rightarrow$ uniform RO-stability}) If a randomized algorithm $\mathcal{A}$ is $\epsilon$-differentially
private, then the  algorithm $\mathcal{A}$ satisfies ($e^{\epsilon}-1$)-RO-stability.
\end{lemma}
%Despite the relation to differential privacy, 
\begin{lemma}
If an algorithm $\mathcal{A}$ is uniform RO-stable with
rate $\epsilon_{stable}(m)$, then $|F_U(\mathcal{A})|$  can
be bounded: $|F_U(\mathcal{A})| \leq \epsilon_{stable}(m)$.
\end{lemma}
Then, we briefly introduce the proof of Theorem 1.
\begin{theorem}
(\textbf{Generalization gap}) If an algorithm $\mathcal{A}$ satisfies $\epsilon$-differential privacy, then the generalization gap can be bounded by
a data-independent constant.
\label{theo:gen_gap}
\end{theorem}
\textbf{Proof:} Given a differentially private algorithm $\mathcal{A}$ with a
privacy cost $\epsilon$, then, according to Lemma 1, we can derive that $\mathcal{A}$ is uniform RO-stability.
Then the generalization bound $|F_U(\mathcal{A})|$ satisfies the following inequality:
\begin{equation}
    |F_U(\mathcal{A})| \leq \epsilon_{stable}
\end{equation}
where $\epsilon_{stable}$ is the stability ratio (equals to $e^{\epsilon}-1$).

\subsection{Proof of Theorem 2}
The proof of Theorem 2 can be accomplished by combing the post-processing
property of the differentially private algorithm and the McDiarmid's inequality.
Thus, we first introduce the post-processing property and the McDiarmid's inequality as follows:
\begin{lemma}
(\textbf{Post-processing}) Let $\mathcal{A}$ be a randomized algorithm that is
$\epsilon$-differently private. Let $f$ be an arbitrary randomized mapping.
Then $f\circ \mathcal{A}$ is $\epsilon$-differentially private.
\end{lemma}
\begin{lemma}
(\textbf{McDiarmid's inequality}) Consider independent random variables
$X_1, X_2, \cdots, X_n \in \mathcal{X}$ and a mapping $f:\mathcal{X}^n\rightarrow
\mathbb{R}$. If, for all $i\in {1,2,\cdots, n}$ and for all $x_1, x_2, \cdots,
x_n, x_i'\in \mathcal{X}$, the function $\phi$ satisfies:
\begin{equation}
    |\phi(x_1, \cdots, x_{i-1}, x_i, x_{i+1}, \cdots, x_n)-\phi(x_1, \cdots, x_{i-1}, x_i', x_{i+1}, \cdots, x_n)|\leq c_i
\end{equation}
then:
\begin{equation}
    P(|\phi(X_1, \cdots, X_n)|-\mathbb{E}\phi|\geq t) \leq 2\exp(-\dfrac{-2t^2}{\sum c_i^2})
\end{equation}

\end{lemma}
\begin{theorem}
(\textbf{Uniform convergence}): Suppose $\mathcal{A}_d$ satisfies $\epsilon$-differential privacy and $\mathbf{d}^{(k)}(x;\theta_{d}^{(k)})$ be the output of $\mathcal{A}_d$ at the $k$-th iteration. Then, $\forall k$, the generalization gap with respect to $\mathbf{d}^{k}$ can be bounded by a universal constant which
is related to $\epsilon$.
\end{theorem}

\noindent \textbf{Proof:} For a training algorithm $\mathcal{A}$ which satisfies $\epsilon$-differentially private,

\noindent then the function build by the output of $\mathcal{A}$ is $\epsilon$-differentially
private in terms of Lemma 3. As a result, the computation of $U(\theta_d, \theta_g^*)$ is
differentially private. Using Lemma 1, we can derive that $U(\theta_d, \theta_g^*)$ 
satisfies uniform RO-stability. 

\noindent We further let $\phi = U$, then based on the stability of $U$, we can meet the requirement of Lemma 4. Based on Lemma 4, we can obtained that:
\begin{equation}
    P(|U(\theta_d, \theta_g^*)-\hat{U}(\theta_d, \theta_g^*)|\geq t) \leq 2\exp(\dfrac{-2t^2}{m\epsilon^2})
\end{equation}
where $\epsilon$ is the privacy cost and m denotes the number of samples used for
computing the empirical loss $\hat{U}$.

\subsection{Hyper-parameter Setting}
Here we list the details of the hyper-parameters of different experiments in Table~\ref{tb:paras}.

\begin{table}[!h]
\label{tb:paras}
\centering
\caption{Details of hype-parameters of Adam optimizer.}
\begin{tabular}{cccc}
\toprule
Strategy                       & Learning rate & $\beta_1$ & $\beta_2$ \\ \hline
\textbf{-JS divergence-}       & \multicolumn{3}{l}{}                  \\ 
Original                       & 0.0004        & 0.5       & 0.999     \\ 
Weight clipping                & 0.0004        & 0.5       & 0.999     \\
Spectral normalization         & 0.0004        & 0.0       & 0.999     \\
Gradient penalty               & 0.0004        & 0.0       & 0.999     \\ \hline
\textbf{-Wassertain Distance-} &               &           &           \\ 
W/o clipping                   & 0.0002        & 0.5       & 0.999     \\ 
Weight clipping                & 0.0002        & 0.5       & 0.999     \\ 
Spectral normalization         & 0.0002        & 0.0       & 0.999     \\
Gradient penalty               & 0.0002        & 0.0       & 0.999     \\ \bottomrule
\end{tabular}
\end{table}
\subsection{Visualization}
Here we show some generated images produced by GANs, which are trained with different regularization techniques. As concluded in the paper, we can see that
spectral normalization achieves comparable quality of generated images in contrast to the original GAN (even better on the LFW dataset). 
\begin{table}[h]
    \centering
    \begin{tabular}{ccc}
        \includegraphics[width=0.3\textwidth]{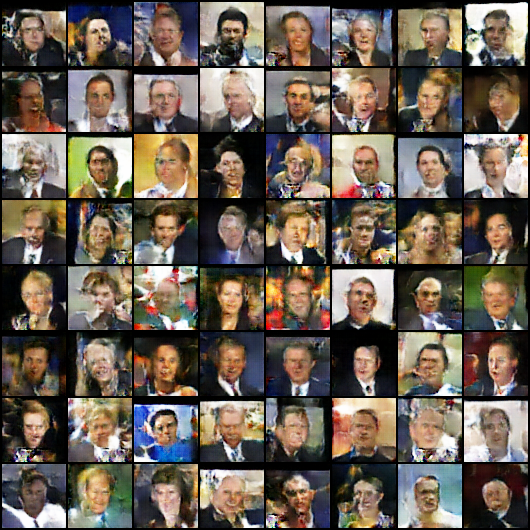}& \includegraphics[width=0.3\textwidth]{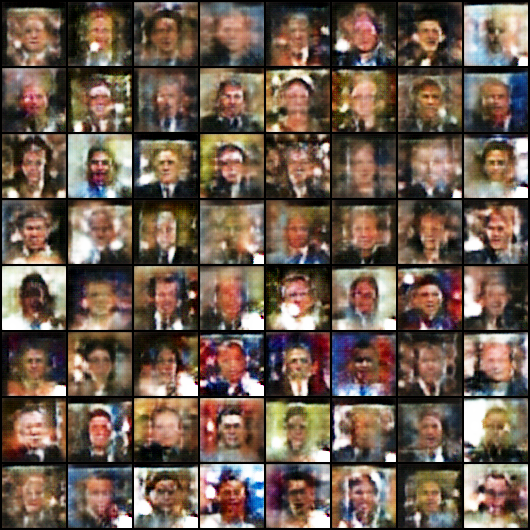}&\includegraphics[width=0.3\textwidth]{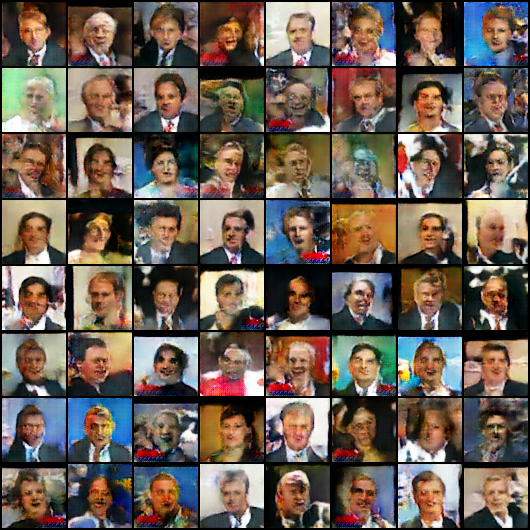}\\
        \includegraphics[width=0.3\textwidth]{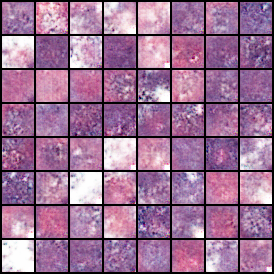}& \includegraphics[width=0.3\textwidth]{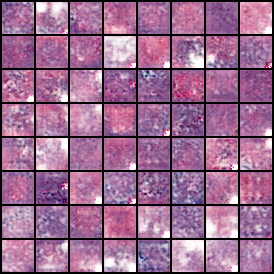}&\includegraphics[width=0.3\textwidth]{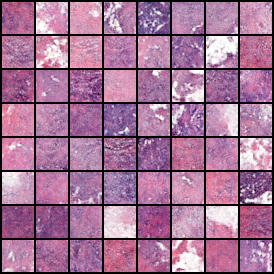}\\
        Original&Clip&Spectral\\
    \end{tabular}
    \caption{Visualization of generated images of GAN models trained with the JS-divergence. The first row is the results of the LFW dataset and the second row is the IDC dataset.}
    \label{tab:my_label}
\end{table}

\end{document}